\documentclass[
]{ceurart}

\usepackage{listings}
\begin{document}

\copyrightyear{2024}
\copyrightclause{Copyright for this paper by its authors.
  Use permitted under Creative Commons License Attribution 4.0
  International (CC BY 4.0).}

\conference{SWAT4HCLS 2024: The 15th International Conference on Semantic Web Applications and Tools for Health Care and Life Sciences}

\title{SPARQL Generation: an analysis on fine-tuning OpenLLaMA for Question Answering over a Life Science Knowledge Graph }


\author[1]{Julio C. {Rangel}}\fnmark[1]
\author[2,3] {Tarcisio {Mendes de Farias}}\fnmark[1]
\author[3] {Ana Claudia Sima}\fnmark[1] 
\author[1]{Norio Kobayashi}\fnmark[1]\cormark[1]

\address[1]{RIKEN Information R\&D and Strategy Headquarters, 2-1 Hirosawa, 351-0198 Wakoshi, Japan}
\address[2]{Department of Ecology and Evolution, University of Lausanne, Switzerland}
\address[3]{SIB Swiss Institute of Bioinformatics, Lausanne, Switzerland}

\cortext[1]{Corresponding author.}
\fntext[1]{All authors contributed equally.}

\begin{abstract}
  The recent success of Large Language Models (LLM) in a wide range of Natural Language Processing applications opens the path towards novel Question Answering Systems over Knowledge Graphs leveraging LLMs. However, one of the main obstacles preventing their implementation is the scarcity of training data for the task of translating questions into corresponding SPARQL queries, particularly in the case of domain-specific KGs. To overcome this challenge, in this study, we evaluate several strategies for fine-tuning the OpenLlama LLM for question answering over life science knowledge graphs. In particular, we propose an end-to-end data augmentation approach for extending a set of existing queries over a given knowledge graph towards a larger dataset of semantically enriched question-to-SPARQL query pairs, enabling fine-tuning even for datasets where these pairs are scarce. In this context, we also investigate the role of semantic "clues" in the queries, such as meaningful variable names and inline comments. Finally, we evaluate our approach over the real-world Bgee gene expression knowledge graph and we show that semantic clues can improve model performance by up to 33\% compared to a baseline with random variable names and no comments included.
\end{abstract}


\maketitle

\section{Introduction}

 Translating natural language questions to SPARQL, a standard query language over RDF knowledge graphs, has been the subject of research for more than a decade \cite{tablan2008natural, sima2021bio}. Recently, the capacity of generalization of large language models (LLMs) to solve different tasks including computer programming has led to breakthroughs in the performance of code generation tasks \cite{haluptzok2023language}. However, although SPARQL queries can be considered as a specialized type of computer code, several studies have pointed out limitations of LLMs for the task of SPARQL query generation \cite{klager2023gpt, sima2023potential}. For instance, generated SPARQL queries can be syntactically correct, however they are semantically wrong. 

  In life sciences, several freely available scientific datasets are accessible through SPARQL endpoints \cite{10.1093/nar/gkad902}. Nevertheless, life scientists are often not able to write SPARQL queries. This is mainly because they do not know SPARQL or how the data are structured. Even for proficient SPARQL users, understanding the data schema of knowledge bases (KB) can be a time-consuming and complex task, which needs to be repeated for every new data source. Applying directly a LLM for Knowledge Graph Question Answering (KGQA) may significantly simplify this process in allowing users to interact with data directly in natural language. However, KGQA systems over scientific data need to demonstrate high accuracy, because researchers cannot base their studies on wrong answers. Therefore, generating SPARQL queries that can be executed over high quality life science datasets can mitigate the problem of providing incorrect answers. This is because answers are generated based on facts stated in a KB rather than directly from the LLM, which is a probabilistic model that can suffer from the problem of hallucinations \cite{sima2023potential}---\textit{i.e.,} generating plausible, but factually incorrect answers. Moreover, scientific KBs are usually curated and high-quality.

To overcome LLM limitations such as hallucinations, we explore several strategies to fine-tune OpenLLaMA, an open source LLM, in order to develop a KGQA system over scientific datasets. These strategies are mainly based on different ways to augment the training set of question-to-SPARQL query pairs and knowledge transfer. As a result, the augmented training set achieves a better coverage of the KB contents.

Finally, we apply our approach on the Bgee gene expression knowledge graph.  
 Bgee is a well-established KB to retrieve and compare gene expression patterns in multiple animal species \citep{bastian2021bgee}. It integrates and harmonises multiple data sources that are based on heterogeneous techniques. The choice of the Bgee data source for applying our methodology is explained in Section \ref{sec:eval}. The main contributions of this article are summarised as follows: (i) a question-to-SPARQL dataset augmentation approach; (ii) a fine-tuned LLM for querying gene expression data;  (iii) a methodology to fine-tune an open LLM for SPARQL query generation over life science knowledge graphs; (iv) a large dataset for scientific question answering over Bgee.

\section{Background and Related Works}

\paragraph{Large Language Models}
The rise of large language models (LLMs) like GPT-3, PaLM, ChatGPT, and LLaMA \citep{brown2020language, chowdhery2022palm, openai2023gpt4, Touvron2023LLaMAOA} has significantly improved the performance of natural language processing (NLP) systems. These models, known for their vast scale and data-intensive training, excel in tasks ranging from mathematical problem-solving to commonsense reasoning \citep{Bubeck2023SparksOA}. Recent research has explored enhancing LLMs through techniques like chain-of-thought prompting (CoT) and instruction tuning \citep{wei_chain--thought_2023, ouyang2022training}. Notably, models like Flan-T5 have achieved superior performance using fewer parameters, benefiting from instruction tuning \citep{chung2022scaling}. Additionally, reinforcement learning from human feedback has shown promise in aligning models with human intentions \citep{ouyang2022training}.

\paragraph{Knowledge Graphs and SPARQL Generation}
Recent advancements in the domain of Large Language Models (LLMs) and knowledge graphs have introduced novel methodologies and frameworks. The Chain of Knowledge (CoK) framework augments LLMs with structured knowledge bases to enhance factual accuracy \citep{li_chain_2023}. Concurrently, the generation of a ``chain of thought" significantly bolsters the reasoning capabilities of LLMs \citep{wei_chain--thought_2023}. In a domain-specific approach, a benchmark dataset for Knowledge Graph Question Answering in Materials Science leverages ChatGPT to translate natural language questions into formal knowledge graph queries \citep{an_knowledge_2023}. Furthermore, controlled natural language has been proposed as a target for KGQA semantic parsing, highlighting the potential of LLMs to parse with reduced training data requirements \citep{gal_language_2023}. The Bio-SODA framework provides a method for natural language processing over structured data, using a graph-based approach to translate user questions into potential SPARQL queries, showing improvements on several real-world datasets~\cite{sima2021bio}. In the broader context of SPARQL generation systems, QAWizard leverages machine learning to learn human experiences in entity type identification and RDF-type identification~\cite{chen_intelligent_2021}. The SGPT approach combines the benefits of end-to-end and modular systems for SPARQL query generation, emphasizing the embedding of linguistic features from questions~\cite{rony_sgpt_2022}. Additionally, the field of neural machine translation for SPARQL query generation has been explored, with comparisons across various models and techniques~\cite{chen_sparql_2021, chen_efficient_2022}.

\paragraph{SPARQL Query Datasets}
The development of benchmark datasets for SPARQL queries, especially in the context of Knowledge Graph Question Answering, has gained significant attention in domain-specific applications. Such benchmarks are instrumental in evaluating the efficacy of models, as demonstrated by efforts that employ models like ChatGPT to seamlessly translate natural language questions into their corresponding technical queries~\cite{an_knowledge_2023}. Recently, the benchmark dataset so-called KQA Pro was released \citep{cao-etal-2022-kqa}. It is a large-scale dataset for complex question answering over a dense subset of the Wikidata\footnote{\url{https://www.wikidata.org}} KB. Wikidata is an open, general-purpose and free KB that is readable and editable by both humans and machines. The Wikidata contents are under a free license (i.e., CC0). Although Wikidata is not a domain specific KB, it contains relevant life science data.


\section{Methodology}
\begin{figure}
  \centering
  \includegraphics[width=\linewidth]{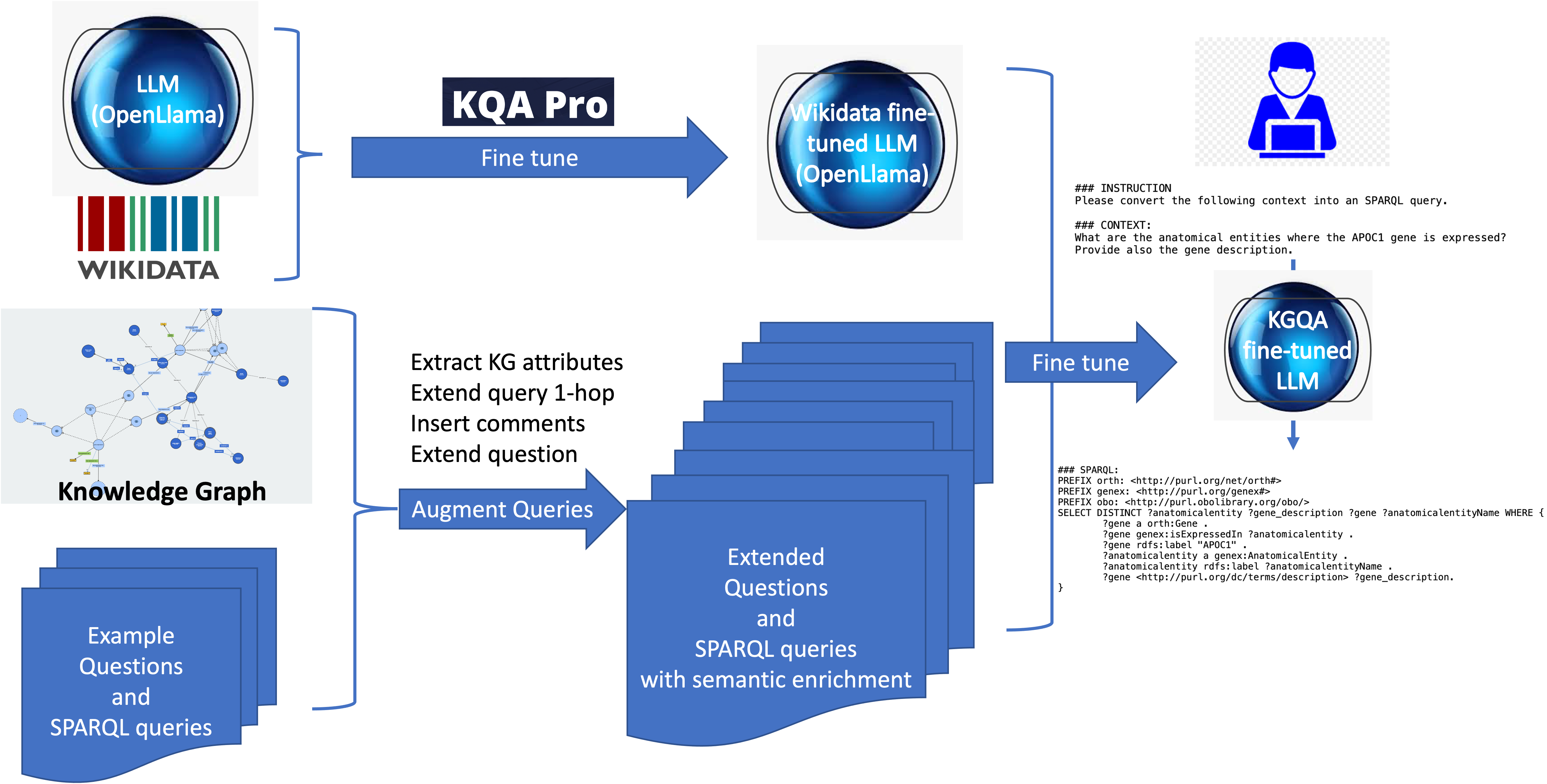}
  \caption{System Architecture. We augment an existing catalog of representative questions over a given knowledge graph and fine-tune OpenLlama in two steps: We first fine-tune the base model using the KQA Pro dataset over Wikidata. Next, we  further fine-tune the resulting model using the extended set of questions and queries over the target knowledge graph. Finally, we obtain a system for Question Answering over Knowledge Graphs (KGQA) which translates natural language user questions into their corresponding SPARQL queries over the target KG.}
  \label{fig:Fig1}
\end{figure}

To produce a more accurate system, our methodology addresses two primary challenges:  question-to-SPARQL query set augmentation from an existing query set; and fine-tuning an open LLM to generate SPARQL queries from plain text questions. An overview of our approach is shown in Figure \ref{fig:Fig1}.

\subsection{Question-to-SPARQL dataset augmentation} 
\label{sec:augm}
Many existing public knowledge graphs provide example queries to guide new users in exploring the available data. However, the number of queries is usually very limited, making them insufficient for fine-tuning LLMs. To mitigate this problem, we designed a dataset augmentation approach to generate extra queries and their corresponding natural language questions, starting from a representative set of existing examples. More precisely, given an example query, we first extract all variables that represent instances of classes in the query. Next, we use both the terminological and assertion axioms of the KG in order to identify datatype properties that can be attached to these instances. We iteratively augment the initial query with 1 extra triple pattern, corresponding to the extra property that can be queried about the class instance. As a concrete example, if an example query contains triple patterns concerning a Gene instance, we additionally generate queries that ask for the gene label, identifier, description and so on. This ensures that the training set will cover a wider range of properties from the KG. 

To investigate the role of the variable names themselves, as well as of additional semantic ``clues" provided in the model prompt (e.g., inline comments providing labels of properties in the query text), we generate the following sets of queries:

\begin{enumerate}
    \item Queries with random variable names and no inline comments.

    For this dataset, we simply remove all comments from the query text and rename all existing variables to random values \textit{?x0}, \textit{?x1} etc.
    \item Queries with meaningful variable names. 
    
    For this dataset, we automatically rename variables with their corresponding class name (e.g. \textit{?gene}, \textit{?anatomicalentity} etc)
    
    \item Queries with meaningful variable names and inline comments

    In addition to meaningful variable names, in this dataset we include inline comments that describe property names (e.g. \textit{?gene obo:RO\_0002162 ?taxon. \# in taxon)}
\end{enumerate}

\subsection{Fine-tuning OpenLLaMA} 

In our training methodology, we employed QLoRA and PEFT to fine-tune the 7 billion parameters version of OpenLlama using four Nvidia A100 40GB GPUs. The OpenLlama fine-tuning was conducted with a learning rate of 2e-05, a batch size of 1, and a maximum context length of 1024, without engaging in extensive hyper-parameter optimization. 

We chose to fine-tune OpenLLaMA\_7b\_v2 with KQA Pro dataset---that contains question-to-SPARQL queries targeting Wikidata. By doing so, we aim to use the fine-tuned model as a base model for translating natural language questions into SPARQL queries over scientific and domain-specific KBs. During the fine-tuning, we applied the Hugging Face SFTTrainer\footnote{\url{https://huggingface.co/docs/trl/sft_trainer}} by completing 13,500 steps on the KQA Pro dataset resulting on the OpenLLaMA+KQA\_Pro model, that is available in our GitHub repository, see Appendix \ref{appendix}.

Finally, we used the Hugging Face SFTTrainer with 2000 steps to further fine-tune OpenLLaMA and OpenLLaMA+KQA\_Pro models with a domain-specific dataset to evaluate our approach as described in Section \ref{sec:eval}.

\section{Evaluation and discussion} 
\label{sec:eval}
We chose Bgee as a scientific KB in the life science domain to evaluate our approach. This is justified mainly by the following reasons: (i) Bgee data are published under a free license (i.e., CC0); (ii) Wikidata contains part of the Bgee data \cite{10.1093/gigascience/giad058}, hence fine-tuning a LLM with Wikidata-related SPARQL query set (i.e., KQA Pro) may help the SPARQL query generation for Bgee; (iii) both of them are real world and large KBs containing billions of triples; (iv) although only about 15 question-to-SPARQL examples are provided by Bgee\footnote{\url{http://purl.org/sib-rdf/bgee-tutorial}}, they are highly representative of its contents according to data providers; (vi) Bgee represents well how complex are the SPARQL queries to answer scientific questions, with queries  composed of multiple triple patterns. The (iv) reason is important for our training set augmentation approach, since we assume that the provided question-to-SPARQL set covers well the main structure of the KB contents. 

To evaluate our methodology, we employed four distinct metrics, each specifically designed for assessing the output of machine translation systems. These metrics were utilized to ascertain the degree of congruence between the SPARQL queries generated by the machine and the corresponding reference queries. The metrics employed in this evaluation include BLEU \cite{goyal-etal-2022-flores}, SP-BLEU \cite{goyal-etal-2022-flores}, METEOR \cite{banerjee2005meteor}, and ROUGE-L \cite{lin2004automatic}. ROUGE-L is an adaptation of the standard ROUGE metric, emphasizing the Longest Common Subsequence, thereby providing insights into sentence-level structural coherence. The BLEU metric quantitatively evaluates the accuracy of word sequences in the output generated by the machine, comparing it to sequences crafted by humans. METEOR extends the capabilities of BLEU by incorporating synonym matching and sentence structural analysis, thus offering a more refined assessment of translation quality.


First, we evaluated the OpenLLaMA\_7b\_v2 model\footnote{\url{https://huggingface.co/openlm-research/open_llama_7b_v2}} without any fine-tuning, that is a zero-shot evaluation, against the Wikidata and Bgee question-to-SPARQL datasets. All computed metrics (i.e., BLEU, SP-BLEU, METEOR, ROUGE-L and F1-score) were either equal or approximately equal to zero. Therefore, the OpenLLaMA\_7b\_v2 model was not capable of generating SPARQL queries that correspond to the datasets of reference without fine-tuning. The OpenLLaMA setting row in Table \ref{tab:bgee} shows in further details the zero-shot evaluation with the Bgee dataset. 

\begin{table}[ht]
\centering
\begin{tabular}{@{}lccccc@{}}
\toprule
Setting & BLEU & SP-BLEU & METEOR & ROUGE-L & F1-score \\ \midrule
OpenLLaMA & 0.00 & 0.00 & 0.006 & 0.014 & 0.005 \\
Bgee random vars & 0.352 & 0.331 & 0.402 & 0.637 & 0.501 \\
Bgee meaningful vars & 0.567 & 0.526 & 0.612 & 0.823 & 0.664 \\
Bgee original & 0.319 & 0.309 & 0.339 & 0.782 & 0.387 \\
Bgee original with comments & 0.474 & 0.442 & 0.520 & 0.826 & 0.570 \\
Bgee meaningful vars comments & 0.555 & 0.526 & 0.593 & 0.845 & 0.639 \\
Wikidata Bgee random vars & 0.261 & 0.237 & 0.291 & 0.642 & 0.392 \\
Wikidata Bgee meaningful vars & 0.382 & 0.347 & 0.448 & 0.666 & 0.531 \\
Wikidata Bgee original & 0.435 & 0.424 & 0.500 & 0.703 & 0.569 \\
Wikidata Bgee meaningful vars comments & \textbf{0.588} & \textbf{0.563} & \textbf{0.628} & \textbf{0.874} & \textbf{0.667} \\
\bottomrule
\end{tabular}
\caption{Performance metrics for the OpenLLama model fine-tuned on the Bgee dataset with various variable naming strategies. Except from the OpenLLaMA setting that means a zero-shot evaluation, the other settings refer to the datasets used to fine-tune the OpenLLaMA model: ``Bgee random vars" contains queries with random variable names and no comments; ``Bgee meaningful vars" is composed of queries with meaningful variable names; ``Bgee original" contains augmented queries without changing variable names and without adding comments; ``Bgee meaningful vars comments" has augmented queries with meaningful variable names and added comments. Settings starting with ``Wikidata" mean the base model is already fine-tuned with a Wikidata query set.}

\label{tab:bgee}
\end{table}

Second, we applied our approach described in Subsection \ref{sec:augm} to generate five different categories of datasets based on the 15 Bgee queries available. Each generated dataset contains 513 queries, hence an increase of about 500 queries per dataset when compared with the original dataset. All generated datasets are available in our GitHub repository (see Appendix \ref{appendix}). These query sets sum into more than 2500 queries over the Bgee KB. The setting column in Table \ref{tab:bgee} shows the five augmented Bgee datasets starting with the ``Bgee" term that were used to fine-tune the OpenLLaMA\_7b\_v2 model. Based on these results, we can conclude that providing meaningful variable names and/or inline comments that define property labels in the SPARQL query significantly improve all evaluated metrics when compared to queries with random variables and without any comment. For instance, the ROUGE-L score improves about 33\% when comparing ``Bgee random vars" versus ``Bgee meaningful vars comments" setting in Table \ref{tab:bgee}.

Third, the results in Table \ref{tab:bgee} demonstrate that using  OpenLLaMA+KQA\_Pro (i.e., rows in Table \ref{tab:bgee} which start with ``Wikidata") as the base model to further fine-tune with the different Bgee query sets does not lead to any significant improvement (or worse, the performance may deteriorate). Nevertheless, these experiments still confirm that providing meaningful variable names and inline comments significantly improve the model. This is clearly noticed with the ``Wikidata Bgee meaningful vars comments" setting that produces the best results for all metrics. Furthermore, if we compare  ``Wikidata Bgee meaningful vars comments" with ``Bgee original" (i.e., query set augmented without adding comments and renaming variables), it indicates an improvement of more than 80\% for the metrics BLEU, SP-BLEU and METEOR. 

\begin{figure}
  \centering
  \includegraphics[width=\linewidth]{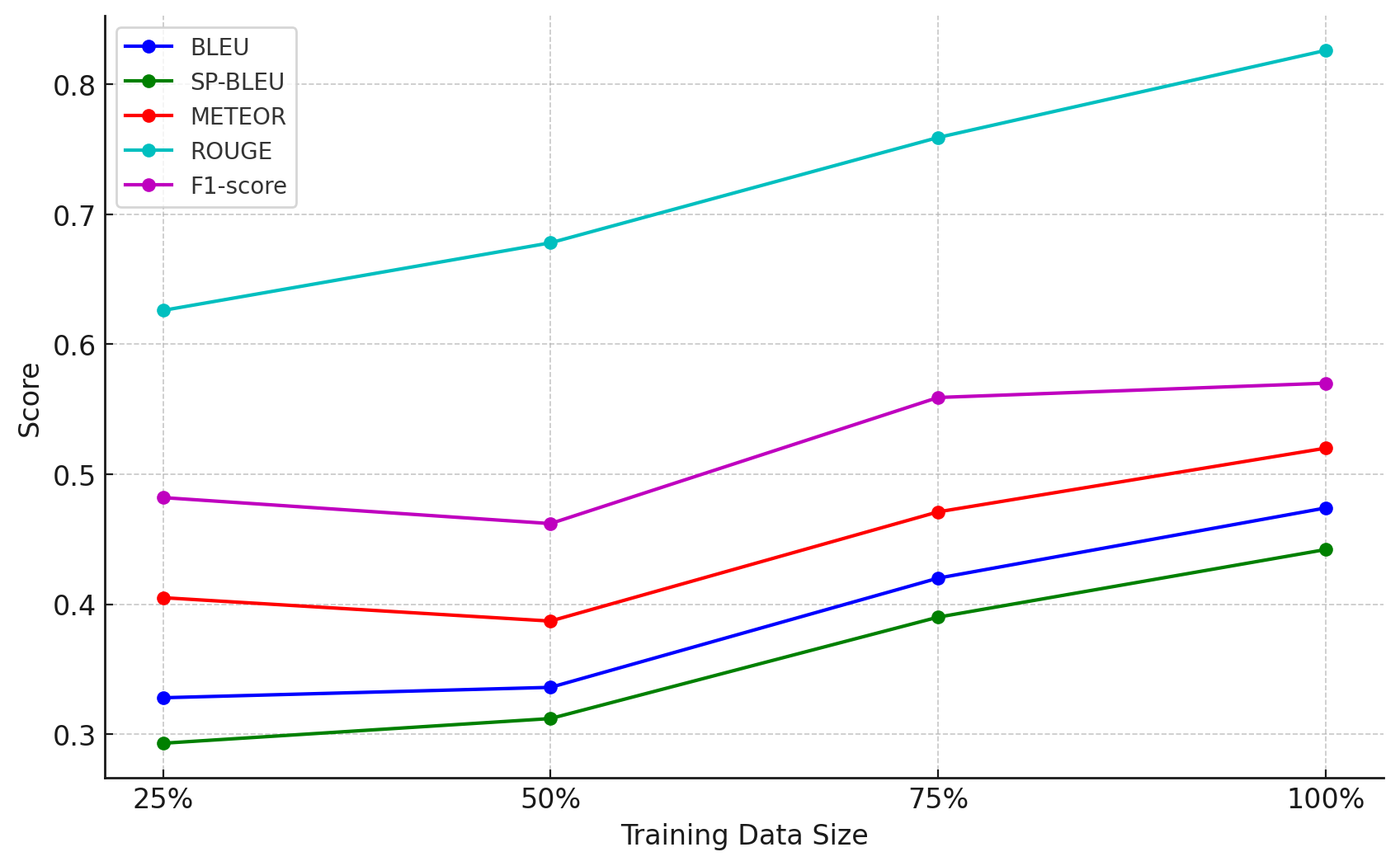}
  \caption{Enhancement in the SPARQL query generation performance (y-axis) with partitions in increments of 25\% of the ``Bgee meaningful vars comments" training set (x-axis).}
  \label{fig:percent}
\end{figure}

Finally, we also evaluated the gain in performance of fine-tuning the model with different augmented query subsets of ``Bgee meaningful vars comments" training set. Figure \ref{fig:percent} illustrates the impact of the dataset augmentation on fine-tuning the OpenLLaMA model for SPARQL query generation within the Bgee dataset context. This figure presents the performance of the model across various metrics at incremental stages of training data augmentation: 25\%, 50\%, 75\%, and 100\% of the training set. The improvement in all metrics with the increase in training data size from 25\% to 100\% suggests that adding more example questions and queries generated automatically through our methodology can help improve the performance of the model in generating more accurate SPARQL queries. 


The different fine-tuned models presented in Table \ref{tab:bgee} for querying Bgee gene expression data are accessible through our GitHub repository, see Appendix \ref{appendix}.
%


\section{Conclusion} 

This study has investigated a methodology for fine-tuning OpenLLaMA to generate SPARQL queries for the task of question answering over a life science knowledge graph. In particular, we have first used KQA\_Pro, a large dataset over Wikidata, to fine-tune the base OpenLLama LLM and then further fine-tuned this model using an augmented dataset of questions and SPARQL queries over the Bgee gene expression database. The following conclusions can be drawn:
\begin{itemize}
    \item Systematically augmenting a representative question-to-SPARQL query set over a scientific KG significantly contributes to improving the performance of the OpenLLaMA model for the SPARQL query generation task. 
    \item Rewriting the SPARQL query to provide more context through inline comments and meaningful variable names considerably improves the OpenLLaMA model. 
    \item The knowledge transfer might deteriorate the LLM performance. Indeed, fine-tuning first OpenLLaMA with an open-domain query set (e.g., KQA Pro) for afterwards fine-tuning it again by targeting a domain-specific KB (e.g., Bgee) can cause the LLM to perform worse than directly fine-tuning the LLM solely with the domain-specific query set.
\end{itemize}

As future work, we plan to improve our query set augmentation approach by also considering property paths of length greater than one (i.e., including the neighbouring instances). Nevertheless, it may require curation since the question-query pairs generated can be nonsensical. Last but not least, we also intend to extend our evaluation to more life science knowledge bases such as the RIKEN metadatabase.



\begin{acknowledgments}
TMF and ACS are thankful for the Swiss Open Research Data Grants (CHORD) in Open Science I, a program coordinated by swissuniversities. TMF thanks also the Canton de Vaud and the SIB Swiss Institute of Bioinformatics---Bgee project.  
\end{acknowledgments}

\bibliography{sample-ceur}

\appendix

\section{Appendix}
\label{appendix}
The materials for this work are available in our GitHub repository at \url{https://github.com/RIKEN-DKO/Generation_SPARQL}.

\end{document}